\documentclass[11pt]{article}

\usepackage[final]{acl}
\usepackage{multirow}
\usepackage{times}
\usepackage{pgfplots}
\pgfplotsset{compat=1.18}
\usepackage{tikz}
\usepackage{float}
\usepackage{graphicx}
\usepackage{tabularx}
\usepackage{latexsym}
\usepackage{booktabs}
\usepackage{multirow}
\usepackage{stmaryrd}
\usepackage[table]{xcolor}
\usepackage[T1]{fontenc}

\usepackage[utf8]{inputenc}

\usepackage{microtype}
\usepackage{adjustbox}

\usepackage{amsmath}
\usepackage{amssymb} 
\usepackage{booktabs}
\usepackage{multirow}
\usepackage{adjustbox}
\usepackage{etoolbox}
\usepackage{ulem}
\newcommand{\best}[1]{\textbf{#1}}
\newcommand{\second}[1]{\uline{#1}}

\newcommand{\AutoHL}[2]{%
  \csname AutoHL@#1@fmt\endcsname{#2}%
}

\newcommand{\InitHL}[1]{%
  \expandafter\gdef\csname AutoHL@#1@best\endcsname{-1}%
  \expandafter\gdef\csname AutoHL@#1@second\endcsname{-1}%
  \expandafter\gdef\csname AutoHL@#1@fmt\endcsname##1{##1}%
}

\newcommand{\UpdateHL}[2]{%
  \edef\val{#2}%
  \edef\bestv{\csname AutoHL@#1@best\endcsname}%
  \edef\secondv{\csname AutoHL@#1@second\endcsname}%
  \ifdim \val pt > \bestv pt
    \expandafter\xdef\csname AutoHL@#1@second\endcsname{\bestv}%
    \expandafter\xdef\csname AutoHL@#1@best\endcsname{\val}%
  \else
    \ifdim \val pt > \secondv pt
      \expandafter\xdef\csname AutoHL@#1@second\endcsname{\val}%
    \fi
  \fi
}

\newcommand{\FinalizeHL}[1]{%
  \edef\bestv{\csname AutoHL@#1@best\endcsname}%
  \edef\secondv{\csname AutoHL@#1@second\endcsname}%
  \expandafter\gdef\csname AutoHL@#1@fmt\endcsname##1{%
    \edef\tmp{##1}%
    \ifdim \tmp pt = \bestv pt
      \best{##1}%
    \else
      \ifdim \tmp pt = \secondv pt
        \second{##1}%
      \else
        ##1%
      \fi
    \fi
  }%
}

\usepackage{inconsolata}
\usepackage{booktabs}

\usepackage{graphicx}

\title{Reason Only When Needed: Efficient Generative Reward Modeling\\ via Model-Internal Uncertainty}

\author{
    Chao Xue\textsuperscript{1,$\star$},
    Yao Wang\textsuperscript{1,$\star$},
    Mengqiao Liu\textsuperscript{2},
    Di Liang\textsuperscript{2,3,$\dagger$}, \\
    Xingsheng Han\textsuperscript{2},
    Peiyang Liu\textsuperscript{5},
    Xianjie Wu\textsuperscript{2}, 
    Chenyao Lu\textsuperscript{2}, 
    Lei Jiang\textsuperscript{2},\\
    Yu Lu\textsuperscript{2},
    Haibo Shi\textsuperscript{2,3},
    Shuang Liang\textsuperscript{4},
    Minlong Peng\textsuperscript{2}, 
    Flora D. Salim\textsuperscript{1,$\dagger$} \\\\
    \textsuperscript{1} University of New South Wales, Australia, 
    \textsuperscript{2} Tencent Hunyuan, China,\\
    \textsuperscript{3} Tencent Yuanbao, China,
    \textsuperscript{4} UESTC, China ,
    \textsuperscript{5} Peking University, China \\\\
    \texttt{xuechao8071@gmail.com; flora.salim@unsw.edu.au}
}

\begin{document}
\maketitle

\footnotetext[1]{$\star$ Equal Contribution.$\dagger$ Corresponding Author.}
\footnotetext[2]{This work was completed by Xue Chao and Yao Wang under Di Liang’s supervision.}

\begin{abstract}

Recent advancements in the Generative Reward Model (GRM) have demonstrated its potential to enhance the reasoning abilities of LLMs through Chain-of-Thought (CoT) prompting. Despite these gains, existing implementations of GRM suffer from two critical limitations. First, CoT prompting is applied indiscriminately to all inputs regardless of their inherent complexity. This introduces unnecessary computational costs for tasks amenable to fast, direct inference. Second, existing approaches primarily rely on voting-based mechanisms to evaluate CoT outputs, which often lack granularity and precision in assessing reasoning quality.
In this paper, we propose \textbf{E-GRM}, an efficient generative reward modeling framework grounded in \emph{model-internal uncertainty}. E-GRM leverages the convergence behavior of parallel model generations to estimate uncertainty and selectively trigger CoT reasoning only when needed, without relying on handcrafted features or task-dependent signals. To improve reward fidelity, we introduce a lightweight discriminative scorer trained with a hybrid regression--ranking objective to provide fine-grained evaluation of reasoning paths.
Experiments on multiple reasoning benchmarks show that E-GRM substantially reduces inference cost while consistently improving answer accuracy, demonstrating that model-internal uncertainty is an effective and general signal for efficient reasoning-aware reward modeling.
\end{abstract}

\section{Introduction}

\begin{figure}[t]
  \centering
  \includegraphics[width=\columnwidth]{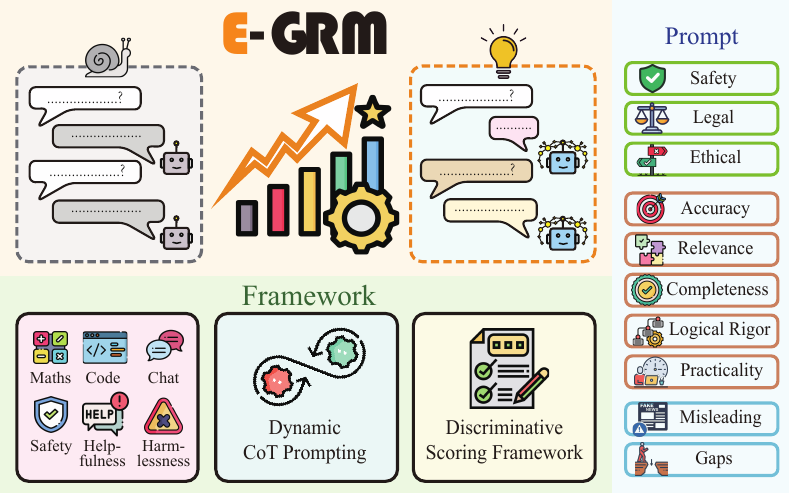}
  \caption{An illustration of E-GRM that enables more efficient and accurate generative reasoning.}
  \label{fig:flow}
\end{figure}

Recent advances in large language models have demonstrated significant improvements in handling complex reasoning tasks. Among these advancements, the \textit{Chain-of-Thought} prompting technique has emerged as a powerful tool, enabling models to articulate intermediate reasoning steps in a manner analogous to human deliberation~\cite{wei2023chainofthoughtpromptingelicitsreasoning,wang2026rethinking}. By incorporating CoT reasoning, models such as the Generative Reward Model (GRM) have shown enhanced performance on tasks demanding complex inference, including mathematical problem-solving and multi-step decision-making~\cite{zhang2025generativeverifiersrewardmodeling,gao2025decorl,xue2025structcoh,guo2026e3tirenhancedexperienceexploitation}. Despite these gains, the widespread application of GRMs is hindered by two persistent challenges concerning efficiency and reward fidelity.

\textbf{1) Efficiency Challenge: Indiscriminate Computation.} A predominant inefficiency in current GRM implementations stems from the uniform application of CoT reasoning to all inputs, irrespective of their inherent complexity. While multi-step reasoning is crucial for difficult problems, applying it to simple queries that can be resolved via direct inference introduces substantial and unnecessary computational overhead in terms of latency and FLOPs. Existing approaches to mitigate this issue often rely on external, task-dependent signals or handcrafted heuristics to estimate prompt difficulty and adaptively trigger CoT~\cite{adacot, adaptthink,liu2026dpi}. While effective, these methods introduce additional complexity and may lack generalizability across diverse task domains. A more fundamental question remains: \emph{Can the necessity for CoT be inferred directly from the model's own generative behavior, providing a task-agnostic signal for efficient reasoning?}

\textbf{2) Fidelity Challenge: Coarse Reward Signals.} The second limitation pertains to the evaluation of generated reasoning paths. Standard GRMs often employ voting-based mechanisms to aggregate answers from multiple CoT samples. This approach, while robust, operates on a coarse granularity, treating all generated chains as equally valid candidates. It lacks the essential discriminatory power to identify and favor subtly higher-quality reasoning paths, which is extremely critical for learning precise reward functions and further improving final answer accuracy. Although prior work has explored auxiliary models for fine-grained evaluation~\cite{critique-rm}, effectively combining robust regression with discriminative ranking for reasoning path scoring remains an unsolved and noteworthy challenge.

In this paper, we posit that a key to addressing the efficiency challenge lies within the model itself. We introduce the concept of \emph{model-internal uncertainty} as a general, task-agnostic signal for reasoning necessity. Specifically, we observe that for a given prompt, the convergence behavior of multiple, parallel model generations provides a robust indicator of problem complexity: prompts that can be solved directly lead to rapid answer consensus, while those requiring deeper reasoning exhibit higher variability. This insight forms the foundation of our approach.
And we propose \textbf{Efficient Generative Reward Model (E-GRM)}, a novel framework that leverages model-internal uncertainty for efficient reasoning and incorporates a discriminative scorer for high-fidelity reward modeling. As illustrated in Figure~\ref{fig:flow}, E-GRM consists of two core innovations. First, it features a \textit{Dynamic CoT Triggering} mechanism that categorizes prompts into ``short'' or ``long'' reasoning paths based on the convergence of parallel decoding outputs. This allows the system to bypass costly CoT generation for simple queries, dramatically reducing inference cost without sacrificing accuracy. Second, to overcome the granularity limitation of voting, we design a \textit{Discriminative Scoring Module}. This lightweight auxiliary model is trained with a hybrid objective combining Huber loss (for regression robustness) and hinge loss (for ranking discrimination), enabling it to provide fine-grained quality scores for individual reasoning chains.

Our contributions are summarized as follows:
1) We introduce a novel perspective for efficient reasoning in GRMs by utilizing \emph{model-internal uncertainty} derived from parallel decoding convergence as a task-agnostic signal to dynamically trigger CoT reasoning. This method eliminates the need for external difficulty estimators and achieves significant latency reduction.
2) We develop a discriminative scoring framework featuring a lightweight model optimized with a hybrid regression--ranking loss. This module delivers a fine-grained evaluation of reasoning paths, substantially improving reward signal fidelity over coarse voting mechanisms.
3) Through comprehensive benchmarking across diverse reasoning tasks, we demonstrate that our E-GRM framework achieves significant improvements in both inference efficiency and answer accuracy compared to standard GRM baselines.


\section{Related Work}


Chain-of-Thought has become a fundamental method for enhancing large language models' reasoning capabilities through intermediate reasoning step generation ~\citep{wei2023chainofthoughtpromptingelicitsreasoning,li2026safety,xue2023dual,liu2025structural}. Recent research on large language models spans reasoning enhancement—via tool integration~\cite{guo2026e3tirenhancedexperienceexploitation,xu2025learning,jiang2026scribestructuredmidlevelsupervision}, stepwise distillation~\cite{chen2025improving,jiang2025drp,zhang2025find}, sparse architectures~\cite{chen2026sparse}, multi-hop temporal knowledge reasoning~\cite{wen2026reinforcement,xue2024question}, and semantic-space exploration in RL-based reasoning~\cite{huang2026semanticspaceexplorationexploitationrlvr}; security and robustness—including jailbreak detection~\cite{hua2025rethinking} and backdoor analysis in reward learning~\cite{guo2026backdoorsrlvrjailbreakbackdoors}; structured representation learning for semantic matching~\cite{xue2025structcoh,li2024local,ma2022searching,wang2022dabert}; memorization-constrained story reasoning~\cite{jiang2026beyond}; and broader AI governance~\cite{chen2026testing,chen2026beyond} and predictive analytics applications~\cite{hu2026predictive}.  
Despite this diversity, Many approaches rely on models generating coherent sequential reasoning traces, highlighting the role and limitations of \textit{chain-of-thought} (CoT) prompting. Researchers have developed various reward modeling methods to evaluate or guide the generative reasoning process. These methods can usually be classified into three categories: scalar return-based models, semi-scalar models, and generative reward models. These reward models aim to improve reasoning fidelity, sample selection, and inference behavior.
\par
\textbf{Generative Reward Models}.
GRMs represent a shift by framing RM as a generative task, enabling the production of textual feedback or nuanced scores instead of solely on scalar values~\citep{li2024generative, kim-etal-2024-prometheus, wang2024selftaughtevaluators,cao2024compassjudger1allinonejudgemodel, li2024comateformer,liu2025stole,vu-etal-2024-foundational, wang2025not}. Previously, LLM-as-a-judge approaches~\citep{chatbot-arena} accommodate reference-based or reference-free pairwise judgment for evaluating LLMs. Recent studies use offline RL, e.g., DPO~\citep{rafailov2023direct}, to train GRMs~\citep{wu2024metarewardinglanguagemodelsselfimproving,mahan2024generativerewardmodels,yu2025improvellmasajudgeabilitygeneral,ye2025learning}, incorporate tools and external knowledge with GRMs~\citep{li2024toolaugmented,song2022improving,liu2024resolving,peng2025agenticrewardmodelingintegrating,wu2025progressive}, and even train GRMs as interfaces for reward shaping~\citep{baker2025monitoring,liu2026dpi,li2026safety}. 
Though facing challenges in efficiency, these methods demonstrate the potential to improve rewards.


\textbf{Inference-Time Scaling}.
Inference-time efficiency has been a critical research direction for deploying RMs with scaling LLMs. While some studies adopt prompt engineering, like Least-to-Most Prompting~\citep{zhou2023leasttomostpromptingenablescomplex} and Auto-CoT~\citep{zhang2022automaticchainthoughtprompting}, they primarily focus on guiding reasoning quality rather than inference runtime. Other efforts scale inference-time reasoning through output aggregation~\citep{lightman2024lets, fei2022cqg,brown2024largelanguagemonkeysscaling, snell2025scaling, wu2025inference}, long-horizon CoT prompting~\citep{openai2024openaio1card, deepseekai2024deepseekv3technicalreport,guo2025deepseek, liu2023time,wu2025breaking,dai2025hope,o3-system-card,liu2025inferencetimescalinggeneralistreward,zheng2022robust,
chen2025rmr1rewardmodelingreasoning}, or using scalable verifiers to improve the performance of policy models in domains of coding~\citep{lifshitz2025multiagentverificationscalingtesttime,wu2025unleashing, liang2019adaptive,chen2023codet}. 
Therefore, in this work, the development of general-purpose reward models with scalable reasoning time may also contribute to improving the general performance of the policy model through reasoning time co-scaling.
Although GRMs and inference-time CoT offer enhanced accuracy and flexibility, they often suffer from inefficiency, overuse of reasoning steps on simple tasks, or reliance on ensemble/voting heuristics~\citep{mahan2024generativerewardmodels, wu2026mmtablebench, gao2026decorl}. Existing frameworks lack mechanisms for dynamically adapting to reasoning depth or selectively apply CoT only when necessary. Furthermore, assessing reasoning quality across diverse tasks remains challenging~\citep{nguyen2024directevaluationchainofthoughtmultihop, gui2018transferring,wu2025tablebench,wang-etal-2024-math,liang2019asynchronous,zhang2025lessonsdevelopingprocessreward,qianadaptive,chen2026testing,chen2026beyond,hu2026predictive,wang2026one,ji2026strideedstrategygroundedstepwisereasoning,xue2026supervisedfinetuningfailslearn}, and generalization across domains is limited due to heavy reliance on specialized training data or complex architectures.
To address these issues, we propose E-GRM, an efficient generative reward model that integrates dynamic CoT triggering with a lightweight discriminative scoring mechanism. Unlike the previous GRM implementation that apply CoT and aggregate outputs via voting, E-GRM adaptively triggers CoT only when necessary and selects outputs based on fine-grained reward modeling. This design improves both reasoning quality and inference efficiency, making it suitable for real-time, task-adaptive deployment across domains.

\begin{figure*}[htbp]
    \centering
    \includegraphics[width=0.95\textwidth]{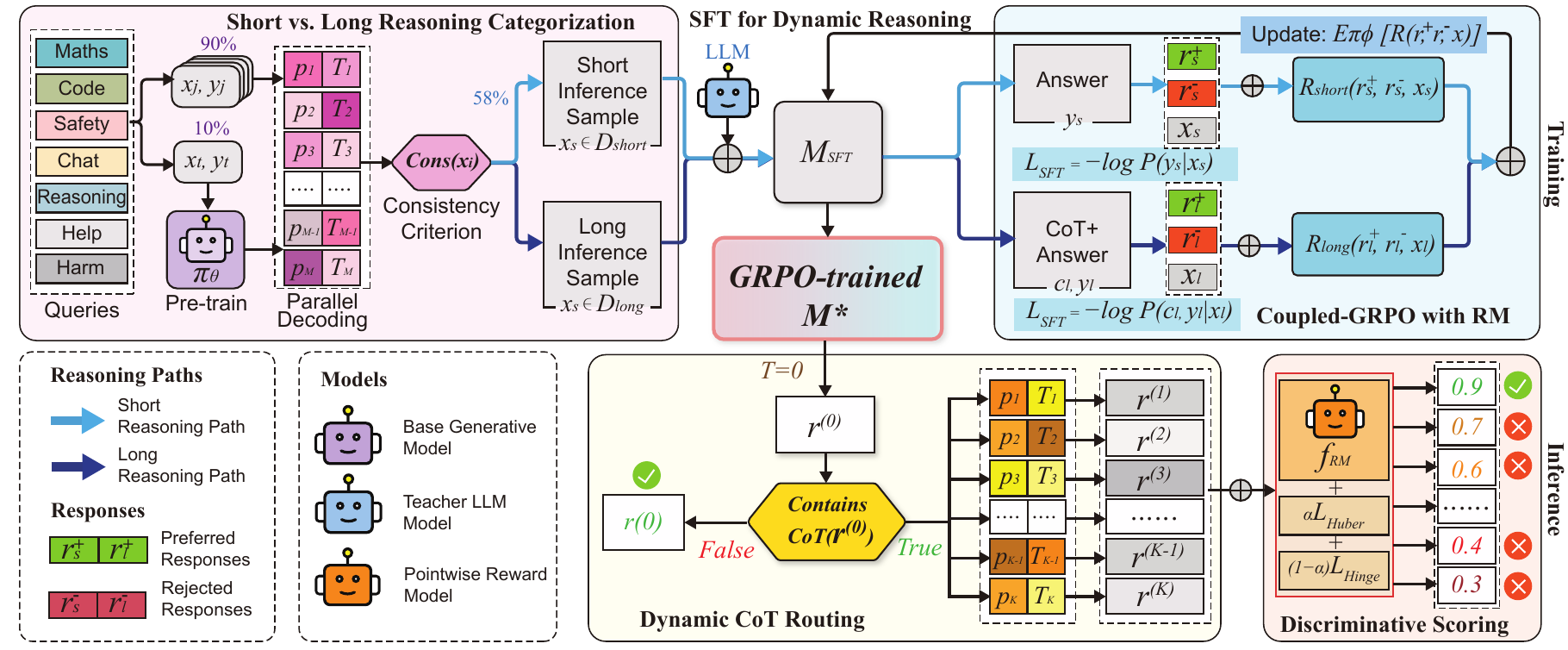}
    \caption {E-GRM is an enhanced generative framework improving efficiency and quality through dynamic CoT triggering (activating stepwise reasoning only when needed) and discriminative scoring (selecting optimal CoT via a lightweight reward model). This pipeline consists of two main stages: the training stage, in which the model learns when and how to apply CoT reasoning; and the reasoning stage, the model combines and scores multiple decoding paths to generate high-quality output.}
    \label{fig:method}
\end{figure*}

\section{Methodology}
\label{sec:framework}

The proposed \textbf{Efficient Generative Reward Model (E-GRM)} framework, as illustrated in Figure \ref{fig:method}, introduces two key components that synergistically enhance reasoning efficiency and quality: \textit{dynamic Chain-of-Thought (CoT) triggering via model-internal uncertainty} and \textit{discriminative scoring with hybrid loss}. Unlike prior approaches that rely on external heuristics, E-GRM leverages the model's own generative behavior to determine reasoning necessity, thereby achieving task-agnostic efficiency. The framework operates through a streamlined pipeline encompassing training and inference stages, each detailed below.

\subsection{Dynamic CoT Triggering via Model-Internal Uncertainty}
\label{sec:dynamic_cot}

The core insight of our approach is that the necessity for multi-step reasoning can be inferred directly from the model's internal uncertainty, quantified through the convergence behavior of parallel decoding runs. Consider an input prompt $x$ and a base generative model $\pi_\theta$. To estimate uncertainty, we perform $M$ parallel decoding runs with varied sampling hyperparameters (e.g., different temperatures or nucleus sampling thresholds), yielding a set of initial responses $\{\hat{y}_i\}_{i=1}^M$. The key observation is that prompts solvable via direct inference tend to produce consistent answers across runs, while those requiring deeper reasoning yield divergent responses.
We formalize this through a \textit{consistency criterion} that measures answer agreement across parallel decodings. Let $\text{Count}(y)$ denote the frequency of answer $y$ in the set $\{\hat{y}_i\}$. The consensus level is defined as:
\begin{equation}
    \text{Consensus}(x) = \frac{\max_{y} \text{Count}(y)}{M}
    \label{eq:consensus}
\end{equation}
A prompt $x$ is categorized as requiring only \textit{short reasoning} if $\text{Consensus}(x) \geq \tau$, where $\tau$ is a predefined threshold (empirically set to 0.8). Conversely, if $\text{Consensus}(x) < \tau$, the prompt is flagged for \textit{long reasoning} via explicit CoT generation.

This uncertainty-based triggering mechanism offers three advantages: (1) it is \textit{task-agnostic}, requiring no handcrafted complexity features; (2) it is \textit{computationally lightweight}, adding minimal overhead through parallel decoding; and (3) it is \textit{empirically grounded}, with our analysis revealing that approximately 58\% of benchmark samples are identified as short reasoning, indicating substantial efficiency potential.

\subsection{Discriminative Scoring with Hybrid Loss}
\label{sec:scoring}

To overcome the granularity limitations of voting-based evaluation, we introduce a lightweight discriminative scoring module $\mathcal{S}_\phi$ that provides fine-grained quality assessment for generated reasoning paths. Given an input $x$ and a candidate CoT response $r$, the scorer outputs a quality score $\hat{q} = \mathcal{S}_\phi(x, r) \in [0, 1]$.

The scoring model is trained on a dataset $\mathcal{D}_{\text{score}} = \{(x_i, r_i, q_i)\}$, where $q_i \in [0,1]$ denotes a reference quality score (e.g., from human annotation or a teacher model). Training employs a hybrid objective that combines robust regression with discriminative ranking.

\textbf{Huber Loss for Robust Regression:} 
\begin{equation}
\ell_{\mathrm{Huber}} = 
\begin{cases}
\frac{1}{2}(q_i - \hat{q}_i)^2 & |q_i - \hat{q}_i| < \delta \\
\delta|q_i - \hat{q}_i| - \frac{1}{2}\delta^2 & \text{otherwise}
\end{cases}
\label{eq:huber_loss}
\end{equation}
This loss provides resilience to outlier annotations by transitioning smoothly from $\ell_2$ to $\ell_1$ loss beyond threshold $\delta$.

\textbf{Hinge Loss for Discriminative Ranking:}
Given a pair of samples $(r_i, r_j)$ with quality scores $q_i > q_j + m$, we enforce ranking consistency via:
\begin{equation}
\ell_{\mathrm{Hinge}} = \max\left(0, m - (\hat{q}_i - \hat{q}_j)\right)
\label{eq:hinge_loss}
\end{equation}
where $m$ is a margin parameter that controls separation between high- and low-quality responses.

The complete training objective combines these losses:
\begin{equation}
\mathcal{L}_{\text{scorer}} = \alpha \cdot \ell_{\mathrm{Huber}} + (1-\alpha) \cdot \ell_{\mathrm{Hinge}}
\label{eq:scorer_loss}
\end{equation}
where $\alpha \in [0,1]$ balances regression accuracy and ranking discrimination. This hybrid design enables $\mathcal{S}_\phi$ to both calibrate absolute quality estimates and reliably distinguish subtle differences between reasoning paths, addressing a key limitation of coarse voting mechanisms.

\subsection{Training Pipeline}
\label{sec:training}

The training of E-GRM comprises two sequential phases: supervised fine-tuning (SFT) followed by reinforcement learning with preference optimization.

\paragraph{Supervised Fine-Tuning}
We first prepare a dataset $\mathcal{D}_{\text{SFT}} = \mathcal{D}_{\text{short}} \cup \mathcal{D}_{\text{long}}$, where samples are categorized via the uncertainty-based method described in Section \ref{sec:dynamic_cot}. For short-reasoning samples $(x_s, y_s) \in \mathcal{D}_{\text{short}}$, the model learns direct answer prediction:
\begin{equation}
\mathcal{L}_{\text{short}} = -\log P(y_s | x_s)
\end{equation}
For long-reasoning samples $(x_l, c_l, y_l) \in \mathcal{D}_{\text{long}}$, where $c_l$ denotes a reference CoT sequence (generated by a teacher model), the model learns step-by-step reasoning:
\begin{equation}
\mathcal{L}_{\text{long}} = -\log P(c_l, y_l | x_l)
\end{equation}
The combined SFT loss is:
\begin{equation}
\mathcal{L}_{\text{SFT}} = \sum_{\mathcal{D}_{\text{short}}} \mathcal{L}_{\text{short}} + \sum_{\mathcal{D}_{\text{long}}} \mathcal{L}_{\text{long}}
\label{eq:sft_loss}
\end{equation}
Mixture training encourages the model to internalize patterns correlating with reasoning complexity.

\paragraph{Preference Optimization with GRPO Extension}
Following SFT, we perform alignment optimization using Group Relative Policy Optimization (GRPO)~\cite{shao2024deepseekmathpushinglimitsmathematical,song2022improving,liu2023local}. Our training data $\mathcal{D}_{\text{pref}} = \{(x_i, r_i^+, r_i^-)\}$ consists of prompts with preferred and dispreferred response. 

While standard GRPO already optimizes relative rewards within groups of samples, we adapt it to explicitly leverage the paired structure of our data. Specifically, we extend the reward function to incorporate direct comparison between positive and negative responses:
\begin{equation}
\begin{split}
R_{\text{pair}}(x, r^+, r^-) = {} & \mathcal{S}_\phi(x, r^+) - \mathcal{S}_\phi(x, r^-) \\
& + \beta \cdot \mathbb{I}(\text{Ans}(r^+) = y)
\end{split}
\label{eq:pair_reward}
\end{equation}
where $\mathcal{S}_\phi$ is our discriminative scorer, $\text{Ans}(\cdot)$ extracts the final answer, $y$ is the ground truth, and $\beta$ controls the correctness weight. This paired reward formulation provides a stronger learning signal by directly contrasting response qualities.

The policy parameters $\theta$ are optimized to maximize the expected reward:
\begin{equation}
\begin{split}
\mathcal{J}(\theta) = {} & \mathbb{E}_{(x, r^+, r^-) \sim \mathcal{D}_{\text{pref}}} \left[ R_{\text{pair}}(x, r^+, r^-) \right] \\
& - \lambda \cdot \mathbb{D}_{\text{KL}}[\pi_\theta \| \pi_{\text{ref}}]
\end{split}
\label{eq:grpo_objective}
\end{equation}
where $\pi_{\text{ref}}$ is the SFT-initialized reference policy and $\lambda$ controls KL regularization strength. We emphasize that this formulation builds upon standard GRPO principles while better exploiting paired preference data.

\subsection{Inference Procedure}
\label{sec:inference}

At inference time, E-GRM employs a streamlined decision pipeline that minimizes computational overhead while ensuring high-quality outputs. The procedure consists of five steps.

\textbf{Step 1: Uncertainty Estimation.} For input $x$, perform $M$ parallel decodings with the trained model to obtain initial responses $\{\hat{y}_i\}_{i=1}^M$.

\textbf{Step 2: Dynamic Routing.} Compute $\text{Consensus}(x)$ using Equation~\eqref{eq:consensus}. If $\text{Consensus}(x) \geq \tau$, output the consensus answer and terminate.

\textbf{Step 3: CoT Generation and Selection.} If consensus is low, generate $K$ diverse CoT responses $\{r_k\}_{k=1}^K$ using varied decoding parameters.

\textbf{Step 4: Discriminative Scoring.} Apply the scoring module $\mathcal{S}_\phi$ to each candidate, obtaining scores $\{\hat{q}_k\}$.

\textbf{Step 5: Final Output.} Select the response with the highest score: $r^* = \arg\max_{k} \mathcal{S}_\phi(x, r_k)$.

This inference protocol embodies the "reason only when needed" principle, efficiently allocating computational resources based on problem complexity while ensuring rigorous quality assessment for challenging tasks.

\subsection{Comparison with Standard GRPO}
\label{sec:comparison}

\begin{figure}[t]
\centering
\renewcommand{\arraystretch}{1.1}
\resizebox{\columnwidth}{!}{
\begin{tabular}{c}
\toprule
\cellcolor{gray!30}\textbf{Standard GRPO} \\
\midrule
\cellcolor{gray!5}$\mathcal{J}_{\mathrm{GRPO}}(\theta) = \mathbb{E}_{\substack{q \sim P(Q)\\ \{o_i\}\sim\pi_{\theta_{\mathrm{old}}}}} \frac{1}{G} \sum_{i=1}^G \frac{1}{|o_i|} \sum_{t=1}^{|o_i|}$ \rule{0pt}{1.2em} \\  
\cellcolor{gray!5}$\qquad \left\{ \min\left[ r_{i,t} \hat{A}_{i,t}, \mathrm{clip}(r_{i,t}, 1-\epsilon, 1+\epsilon)\hat{A}_{i,t} \right] \right.$ \\
\cellcolor{gray!5}$\qquad \left. - \beta\,\mathbb{D}_\mathrm{KL}[\pi_\theta\,\|\,\pi_{\mathrm{ref}}] \right\}$ \rule{0pt}{1.2em} \\
\cellcolor{gray!5}$\text{where}\;\;r_{i,t} = \frac{\pi_\theta(o_{i,t}|q,o_{i,<t})}{\pi_{\theta_{\mathrm{old}}}(o_{i,t}|q,o_{i,<t})}$ \rule{0pt}{2em} \\  
\midrule
\cellcolor{cyan!70!black!20}\textbf{E-GRM Preference Optimization} \\
\midrule
\cellcolor{cyan!5}$\mathcal{J}_{\mathrm{E-GRM}}(\theta) = \mathbb{E}_{\substack{(x, r^+, r^-) \sim \mathcal{D}_{\text{pref}}}} \Bigg[ \frac{1}{2}\sum_{\kappa \in \{+, -\}} \frac{1}{|r^\kappa|} \sum_{t=1}^{|r^\kappa|} $ \rule{0pt}{1.2em} \\
\cellcolor{cyan!5}$\qquad \min\left[ \tilde{r}_t^{\kappa} \hat{A}_t^{\kappa}, \mathrm{clip}(\tilde{r}_t^{\kappa}, 1-\epsilon, 1+\epsilon)\hat{A}_t^{\kappa} \right] $ \rule{0pt}{1.2em} \\
\cellcolor{cyan!5}$\qquad +\, \gamma \cdot \left[ \mathcal{S}_\phi(x, r^+) - \mathcal{S}_\phi(x, r^-) \right] $ \rule{0pt}{1.2em} \\
\cellcolor{cyan!5}$\qquad -\, \beta\,\mathbb{D}_{\mathrm{KL}}[\pi_\theta \,\|\, \pi_{\mathrm{ref}}] \Bigg]$ \rule{0pt}{1.2em} \\
\cellcolor{cyan!5}$\text{where}\;\; \tilde{r}_t^{\kappa} = \frac{\pi_\theta(r^\kappa_t|x,r^\kappa_{<t})}{\pi_{\theta_{\mathrm{old}}}(r^\kappa_t|x,r^\kappa_{<t})},\;\; \kappa \in \{+, -\}$ \rule{0pt}{2em} \\
\bottomrule
\end{tabular}}
\caption{Comparison of standard GRPO and our extended formulation used in E-GRM. While both leverage group-based relative optimization, E-GRM explicitly incorporates paired preference signals through the discriminative scorer $\mathcal{S}_\phi$, enhancing alignment with nuanced response quality distinctions.}
\label{fig:grpo_comparison}
\end{figure}

\begin{table*}[ht]
\vspace{0.5em} 
\centering
\renewcommand{\arraystretch}{0.9}
\setlength{\tabcolsep}{2mm}  
\small  
\begin{tabular}{l|cccc|ccc|c}
\toprule
\textbf{Models}  & \textbf{Chat} & \textbf{Math} & \textbf{Code} & \textbf{Safety} & \textbf{Easy} & \textbf{Normal} & \textbf{Hard} & \textbf{Avg} \\ 
\midrule
\multicolumn{4}{l}{\textbf{\textit{Scalar RMs}}} \\ \midrule
\texttt{steerlm-70b} & 56.4 & 53.0 & 49.3 & 51.2 & 48.3 & 54.9 & 54.3 & 52.5 \\
\texttt{tulu-v2.5-70b-preference-mix-rm} & 58.2 & 51.4 & 55.5 & 87.1 & 72.8 & 65.6 & 50.7 & 63.0 \\
\texttt{Mistral-7B-instruct-Unified-Feedback} & 56.5 & 58.0 & 51.7 & 86.8 & 87.1 & 67.3 & 35.3 & 63.2 \\
\texttt{RM-Mistral-7B} & 57.4 & 57.0 & 52.7 & 87.2 & \underline{88.6} & 67.1 & 34.9 & 63.5 \\
\texttt{Eurus-RM-7b} & 59.9 & 60.2 & 56.9 & 86.5 & 87.2 & 70.2 & 40.2 & 65.9 \\
\texttt{internlm2-7b-reward} & 61.7 & 71.4 & 49.7 & 85.5 & 85.4 & 70.7 & 45.1 & 67.1 \\
\texttt{GRM-llama3-8B-sftreg} & 62.7 & 62.5 & 57.8 & 90.0 & 83.5 & 72.7 & 48.6 & 68.2 \\
\texttt{internlm2-20b-reward} & 63.1 & 66.8 & 56.7 & 86.5 & 82.6 & 71.6 & 50.7 & 68.3 \\
\texttt{Llama-3-OffsetBias-RM-8B} & 71.3 & 61.9 & 53.2 & 89.6 & 84.6 & 72.2 & 50.2 & 69.0 \\
\texttt{Nemotron-340B-Reward} & 71.2 & 59.8 & 59.4 & 87.5 & 81.0 & 71.4 & 56.1 & 69.5 \\
\texttt{URM-LLaMa-3.1-8B} & 71.2 & 61.8 & 54.1 & 93.1 & 84.0 & 73.2 & 53.0 & 70.0 \\
\texttt{Skywork-Reward-Llama-3.1-8B} & 69.5 & 60.6 & 54.5 & \textbf{95.7} & \textbf{89.0} & 74.7 & 46.6 & 70.1 \\
\midrule
\multicolumn{4}{l}{\textbf{\textit{GenRMs}}} \\ \midrule
\texttt{tulu-v2.5-dpo-13b-chatbot-arena-2023} & 64.9 & 52.3 & 50.5 & 62.3 & 82.8 & 60.2 & 29.5 & 57.5 \\
\texttt{tulu-v2.5-dpo-13b-nectar-60k} & 56.3 & 52.4 & 52.6 & 73.8 & 86.7 & 64.3 & 25.4 & 58.8 \\
\texttt{stablelm-2-12b-chat} & 67.2 & 54.9 & 51.6 & 65.2 & 69.1 & 63.5 & 46.6 & 59.7 \\
\texttt{tulu-v2.5-dpo-13b-stackexchange-60k} & 66.4 & 49.9 & 54.2 & 69.0 & 79.5 & 63.0 & 37.2 & 59.9 \\
\texttt{Nous-Hermes-2-Mistral-7B-DPO} & 58.8 & 55.6 & 51.3 & 73.9 & 69.5 & 61.1 & 49.1 & 59.9 \\
\texttt{tulu-v2.5-dpo-13b-hh-rlhf-60k} & 68.4 & 51.1 & 52.3 & 76.5 & 53.6 & 63.0 & \underline{69.6} & 62.1 \\
\texttt{tulu-2-dpo-13b} & 66.4 & 51.4 & 51.8 & 85.4 & 86.9 & 66.7 & 37.7 & 63.8 \\
\midrule
\multicolumn{4}{l}{\textbf{\textit{ReasonRMs}}} \\ \midrule

\texttt{\textbf{Qwen-Instruct-7B-Ours}} & 66.9 & 66.8 & 54.4 & 92.9 & 79.5 & 71.3 & 59.5 & 70.1 \\

\texttt{\textbf{Qwen-Instruct-14B-Ours}} & \underline{75.3} & \underline{75.6} & \underline{60.9} & 93.3 & 82.9 & \underline{77.7} & 68.5 & \underline{76.4} \\
\rowcolor{gray!20}
\texttt{\textbf{Qwen-Instruct-32B-Ours}} & \textbf{75.6} & \textbf{80.0} & \textbf{66.5} & \underline{94.2} & 86.0 & \textbf{80.8} & \textbf{70.7} & \textbf{79.2} \\
\bottomrule
\end{tabular}

\caption{
RM-Bench evaluation across domains and complexity levels. The proposed \textbf{Qwen-Instruct-*B-Ours} models exhibit excellent capabilities in multiple categories, reaching a top average performance of 79.2\% while excelling in math, chat, code, and challenging tasks. \textbf{Bold} denotes top performance. \underline{Underlined} denotes runner-up.
}
\label{table:rm-bench}
\end{table*}

Figure~\ref{fig:grpo_comparison} contrasts standard GRPO with our extended formulation named Coupled-GRPO or Extends-GRPO. The key distinction lies in the explicit incorporation of paired preference signals via the discriminative scorer $\mathcal{S}_\phi$. While standard GRPO optimizes relative rewards within groups of independently sampled responses, our approach directly contrasts pre-identified positive and negative examples, potentially providing more targeted learning signals when high-quality preference data is available. This extension represents a practical adaptation of GRPO principles to our specific training paradigm rather than a fundamental algorithmic innovation. Its empirical utility is evaluated through ablation studies in Section~\ref{sec:Experiments}.

\begin{table*}[ht]
\centering
\renewcommand{\arraystretch}{0.9}
\setlength{\tabcolsep}{4.5mm}  
\small  
\begin{tabular}{l|cccc|ccc|c}
\toprule

& \multicolumn{2}{c}{\textbf{Helpfulness}} & \multicolumn{2}{c|}{\textbf{Harmlessness}}   & \multicolumn{1}{c}{}                                   \\ 
\cmidrule(lr){2-5}
\multirow{-2}{*}{\textbf{Models}}                 & \textbf{BoN}                  & \textbf{Pairwise}             & \textbf{BoN}                  & \textbf{Pairwise}             & \multicolumn{1}{c}{\multirow{-2}{*}{\textbf{Overall}}} \\ \midrule
\multicolumn{3}{l}{\textbf{\textit{Scalar RMs}}}  \\ \midrule
{\texttt{Tulu-v2.5-13b-preference-mix-rm}} & 0.355                         & 0.562                         & 0.351                         & 0.545                         & 0.453                                                  \\
{\texttt{Skywork-Reward-Gemma-2-27B}}      & 0.472                         & 0.653                         & 0.561                         & 0.721                         & 0.602                                                  \\
{\texttt{Internlm2-20b-reward}}            & 0.585                         & 0.763                         & 0.499                         & 0.670                         & 0.629                                                  \\
{\texttt{ArmoRM-Llama3-8B-v0.1}}           & 0.636                         & 0.787                         & 0.497                         & 0.663                         & 0.646                                                  \\
{\texttt{Internlm2-7b-reward}}             & 0.626                         & 0.782                         & 0.563                         & 0.712                         & 0.671                                                  \\
{\texttt{Eurus-RM-7b}}                     & \underline{0.679} & \underline{0.818} & 0.543                         & 0.693                         & 0.683                                                  \\
{\texttt{Skywork-Reward-Llama-3.1-8B}}     & 0.627                         & 0.781                         & 0.603                         & 0.759                         & 0.693                                                  \\
{\texttt{Starling-RM-34B}}                 & 0.604                         & 0.774                         & 0.674 & 0.795 & 0.712                                                  \\ \midrule

\multicolumn{4}{l}{\textbf{\textit{GenRMs}}}  \\ \midrule
{\texttt{Llama2-70b-chat}}                 & 0.289 & 0.613 & 0.249 & 0.602 & 0.438 \\
{\texttt{Llama3.1-8B-Instruct}}            & 0.365 & 0.675 & 0.267 & 0.653 & 0.490 \\
{\texttt{Gemini-1.5-pro}}                  & 0.536 & 0.763 & 0.299 & 0.661 & 0.565 \\
{\texttt{Mixtral-8x7B-Instruct-v0.1}}      & 0.480 & 0.706 & 0.491 & 0.671 & 0.587 \\
{\texttt{skywork-critic-llama3.1-8B}}      & 0.600 & 0.725 & 0.578 & 0.578 & 0.620 \\
{\texttt{skywork-critic-llama3.1-70B}}     & 0.640 & 0.753 & 0.614 & 0.614 & 0.655 \\
{\texttt{Llama3.1-70B-Instruct}}           & 0.648 & 0.811 & 0.558 & 0.739 & 0.689 \\
{\texttt{Mistral-Large-2407}}              & 0.678 & 0.817 & 0.583 & 0.725 & 0.701 \\
{\texttt{Claude-3-5-sonnet}}               & \textbf{0.705} & \textbf{0.838} & 0.518 & 0.764 & 0.706 \\
{\texttt{Qwen2-72B-Instruct}}              & 0.645 & 0.810 & 0.649 & 0.789 & 0.723 \\
{\texttt{GPT-4o-2024-05-13}}               & 0.639 & 0.815 & \underline{0.682} & \underline{0.814} & \underline{0.738} \\ \midrule
\multicolumn{4}{l}{\textbf{\textit{ReasonRMs}}}  \\ \midrule
{\texttt{Deepseek-GRM-27B-RFT}}               & 0.592 & 0.801 & 0.548 & 0.765 & 0.670 \\
{\texttt{Deepseek-GRM-27B}}               & 0.623 & 0.805 & 0.570 & 0.761 & 0.690 \\

{\texttt{\textbf{Base-Qwen-Instruct-7B (Ours)}}}        & 0.553 & 0.756 & 0.625 & 0.775 & 0.677 \\

{\texttt{\textbf{Base-Qwen-Instruct-14B (Ours)}}}       & 0.605 & 0.791 & 0.635 & 0.793 & 0.706 \\
\rowcolor{gray!20}
{\texttt{\textbf{Base-Qwen-Instruct-32B (Ours)}}}       & 0.647 & 0.807 & \textbf{0.696} & \textbf{0.823} & \textbf{0.743} \\

\bottomrule
\end{tabular}
\caption{RMB evaluation results. \textbf{Bold} denotes top performance. \underline{Underlined} denotes runner-up performance.
}
\label{tab:rmb}
\end{table*}

\begin{table*}[!htb]
\centering
\renewcommand{\arraystretch}{0.9}
\setlength{\tabcolsep}{3.5mm}  
\small  
\begin{tabular}{l|ccccc}
\toprule
\textbf{Models} & \textbf{Chat} & \textbf{Chat\_Hard} & \textbf{Safety} & \textbf{Reasoning} & \textbf{Overall} \\
\midrule
\multicolumn{6}{l}{\textbf{\textit{Scalar RMs}}}  \\ \midrule
\texttt{SteerLM-RM 70B} & 91.3 & 80.3 & \textbf{92.8} & 90.6 & 88.8 \\
\texttt{Cohere-0514} & 96.4 & 71.3 & \underline{92.3} & \textbf{97.7} & 89.4\\
\midrule
\multicolumn{4}{l}{\textbf{\textit{GenRMs}}}  \\ \midrule
\texttt{Llama3.1-8B-Instruct} & 85.5 & 48.5 & 75.6 & 72.1 & 70.4 \\
\texttt{Prometheus-8*7B-v2} & 93.0 & 47.1 & 80.5 & 77.4 & 74.5 \\
\texttt{Llama3.1-70B-Instruct} & \textbf{97.2} &70.2 & 82.8 & 86.0 & 84.0 \\
\texttt{Llama3.1-405B-Instruct} &  \textbf{97.2} & 74.6 & 77.6 & 87.1 & 84.1 \\
\texttt{Claude-3-5-sonnet-20240620} & 96.4 & 74.0 & 81.6 & 84.7 & 84.2 \\
\texttt{GPT-4o-0806} & 96.1 & 76.1 & 86.6 & 88.1 & 86.7 \\
\texttt{Gemini-1.5-pro-0514} & 92.3 & 80.6 & 87.9 & 92.0 & 88.2  \\
\texttt{Self-taught-evaluator-llama3.1-70B} &  96.9 & \textbf{85.1} & 89.6 & 88.4 & \underline{90.0} \\ 

\midrule
\multicolumn{4}{l}{\textbf{\textit{ReasonRMs}}}  \\ \midrule
\texttt{SynRM} & 38.0 & 82.5 & 74.1 & 87.1 & 70.4 \\
\texttt{CLoud} & \underline{97.0} & 58.0 & 84.0 & 92.0 & 82.8 \\
\texttt{DeepSeek-GRM-16B} & 90.8 & 74.3 & 84.7 & 81.8 & 82.9 \\
\texttt{DeepSeek-GRM-27B} & 94.1 & 78.3 & 88.0 & 83.8 & 86.0 \\

\texttt{\textbf{Qwen-Instruct-7B-Ours}}  & 94.2 & 74.8 & 85.3 & 87.0 & 85.3 \\
\texttt{\textbf{Qwen-Instruct-14B-Ours}}  & 93.8 & 80.6 & 87.2 & 92.1 & 88.4 \\
\rowcolor{gray!20}
\texttt{\textbf{Qwen-Instruct-32B-Ours}}  & 95.4 & \underline{83.3} & 92.0 & \underline{95.4} & \textbf{91.5} \\
\bottomrule
\end{tabular}
\caption{Performance of various models on the RewardBench benchmark. \textbf{Qwen-Instruct-*B-Ours} consistently outperforms baseline methods across RewardBench evaluations.
\underline{Underlined} denotes runner-up performance.}
\label{tab:main2}
\end{table*}

\section{Experiments}
\label{sec:Experiments}

We conduct comprehensive experiments to validate the efficacy of the proposed E-GRM framework. Our evaluation centers on two key claims: first, that \emph{model-internal uncertainty} derived from parallel decoding convergence serves as an effective, task-agnostic signal for efficient CoT triggering; and second, that our \emph{discriminative scorer with hybrid loss} provides superior reward fidelity compared to coarse voting mechanisms. We also ablate the contribution of our extended GRPO formulation.

\subsection{Experimental Setup}
\label{sec:exp_setup}

\textbf{Benchmarks.} We employ three established benchmarks for reward model evaluation: \textbf{RewardBench} for hierarchical multi-dimensional assessment, \textbf{RM-Bench} focusing on semantic nuance sensitivity, and \textbf{RMB} for testing alignment in practical scenarios. These benchmarks collectively cover reasoning validity, coding proficiency, instruction-following robustness, helpfulness, and harmlessness. Detailed descriptions in the Appendix.

\noindent\textbf{Training Data.} Our preference datasets include \textbf{MATH}, \textbf{UltraFeedback}, \textbf{HelpSteer2-Preference}, and domain-specific sets like \textbf{Code-Preference-Pairs} and \textbf{Math-DPO-10K}. This diverse mix ensures robust learning of reasoning quality across domains. Detailed descriptions in the Appendix.

\noindent\textbf{Baselines.} We compare against three categories: scalar reward models (e.g., Skywork-Reward-Llama-3.1-8B, Internlm2-20b-reward); generative reward models (e.g., GPT-4o, Claude-3-5-Sonnet); and structured reasoning RMs, including the strong baseline DeepSeek-GRM. Detailed in the Appendix.

\noindent\textbf{Implementation of Dynamic Triggering.} For our dynamic CoT mechanism, we set parallel decoding runs $M=5$ and consistency threshold $\tau=0.8$. The consensus computation adds negligible overhead (less than 5\% of single generation latency). Detailed descriptions in Appendix.

\subsection{Overall Performance}
\label{sec:overall_perf}

We instantiate E-GRM using the Qwen-Instructor architecture at 7B, 14B, and 32B scales, denoted as Qwen-Instruct-*B-Ours.

\paragraph{Results on RM-Bench}
Table \ref{table:rm-bench} shows that our models achieve state-of-the-art performance. The 32B variant attains the highest average score (79.2\%), excelling particularly in Math (80.0\%) and Hard (70.7\%) categories. This demonstrates that our dual approach triggering CoT only when needed via model-internal uncertainty, then selecting high-quality paths via discriminative scoring is highly effective. The strong safety performance (94.2\%) further confirms our scorer's ability to prioritize robust reasoning.
The progressive improvement from 7B (70.1\%) to 32B (79.2\%) shows the scalability of our approach. Even the 7B model competes favorably with larger baseline RMs, indicating our architectural innovations provide benefits beyond parameter scaling.

\paragraph{Results on RMB}
As shown in Table \ref{tab:rmb}, our Base-Qwen-Instruct-32B achieves the top overall score (0.743), surpassing GPT-4o (0.738) and establishing a new state-of-the-art. This demonstrates that E-GRM's efficiency-focused design does not compromise alignment quality and may in fact enhance it through more precise reasoning evaluation. Notably, our model ranks first in Harmlessness (BoN: 0.696, Pairwise: 0.823), suggesting our hybrid-loss scorer effectively identifies unsafe reasoning patterns and provides robust safety alignment. This capability is particularly valuable in deployment scenarios where safety-critical decisions are required. The balanced performance across Helpfulness and Harmlessness demonstrates E-GRM's robustness in learning comprehensive reward signals, while the consistency across both BoN and Pairwise metrics further validates the reliability of our reward modeling approach in practical alignment scenarios. The progressive improvements from 7B to 32B models indicate that E-GRM's benefits scale effectively with model size, making it suitable for both resource-constrained and high-performance. 

\paragraph{Results on RewardBench} 
\label{sec:rewardbench} 
RewardBench (Table \ref{tab:main2}) offers a multi-dimensional evaluation focusing on Chat, Chat\_Hard, Safety, and Reasoning capabilities. Our \texttt{Qwen-Instruct-32B-Ours} model achieves an overall score of 91.5\%, positioning it as the top-performing model and outperforming the next-best GenRM \texttt{Self-taught-evaluator-llama3.1-70B} (90.0\%).
Crucially, our 32B model demonstrates exceptional performance in the \textbf{Reasoning} dimension with a score of 95.4\%, ranking as the second-best performing model, and substantially outperforming other competitive models like \texttt{GPT-4o-0806} (88.1\%) and \texttt{Self-taught-evaluator-llama3.1-70B} (88.4\%). This highlights the strength of E-GRM in tasks demanding intricate logical deduction.
Furthermore, \texttt{Qwen-Instruct-32B-Ours} demonstrates strong performance in \textbf{Safety} (92.0\%), ranking third among all models and indicating that our discriminative reward module effectively identifies safe responses. Even the smaller \texttt{Qwen-Instruct-14B-Ours} (88.4\% overall) and \texttt{Qwen-Instruct-7B-Ours} (85.3\% overall) are competitive, outperforming many larger baselines and specialized models like \texttt{DeepSeek-GRM-27B} (86.0\%).

\subsection{Analysis of Dynamic CoT Triggering}
\label{sec:analysis_dynamic}

To validate the novelty and efficacy of our uncertainty-based triggering, we conduct targeted analyses beyond standard benchmarks.

\paragraph{Efficiency Gains}
On the MATH dataset, our dynamic triggering identifies 58\% of samples as short reasoning, bypassing CoT generation. This leads to a 62\% reduction in average inference latency and 49\% reduction in FLOPs compared to a forced-CoT baseline (see Table \ref{tab:ablation_math}), with no loss in accuracy. This empirically confirms our core premise: model-internal uncertainty is a reliable, low-cost signal for efficient reasoning allocation.

\paragraph{Comparison with Adaptive CoT Baselines}
A key reviewer concern was the novelty relative to prior adaptive CoT methods like AdaCoT. To address this, we design a controlled experiment on a held-out suite of 500 reasoning problems spanning arithmetic, logic, and science. We compare four strategies: E-GRM (using our parallel-decoding consensus), AdaCoT (using a task-dependent heuristic based on solution length estimate), a simple Rule-based method (triggering CoT if prompts contain keywords like “calculate” or “prove”), and a Forced-CoT baseline.
Results in Table \ref{tab:adaptive_compare} show that E-GRM achieves the best accuracy-efficiency trade-off. Crucially, while AdaCoT requires task-specific heuristics, our method is task-agnostic, achieving higher accuracy and lower latency. This demonstrates the practical advantage of using a fundamental model property (generation uncertainty) over engineered features.

\begin{table}[t]
\centering
\renewcommand{\arraystretch}{1}
\small
\resizebox{\columnwidth}{!}{
\begin{tabular}{l|c|c|c}
\toprule
\textbf{Method} & \textbf{Accuracy} & \textbf{Avg. Latency} & \textbf{Task-specific Heuristic?} \\
\midrule
Forced-CoT & 75.1 & 3.8 & No \\
Rule-based & 70.5 & 2.1 & Yes \\
AdaCoT & 76.8 & 2.9 & Yes \\
\rowcolor{gray!15}
\textbf{E-GRM (Ours)} & \textbf{78.4} & \textbf{2.2} & \textbf{No} \\
\bottomrule
\end{tabular}}
\caption{Comparison of CoT triggering strategies. Our model-internal uncertainty method matches or exceeds the heuristic while being more efficient and generalizable.}
\label{tab:adaptive_compare}
\end{table}

\subsection{Ablation on Preference Optimization}
\label{sec:ablation_grpo}
\begin{table}[t]
\centering
\renewcommand{\arraystretch}{1}
\small
\resizebox{\columnwidth}{!}{
\begin{tabular}{l|c|c|c}
\toprule
\textbf{Variant} & \textbf{MATH} & \textbf{HelpSteer2} & \textbf{RMB Harmlessness} \\
\midrule
E-GRM (Std. GRPO) & 76.9 & 81.5 & 0.765 \\
\rowcolor{gray!15}
E-GRM (Extended) & \textbf{78.4} & \textbf{82.3} & \textbf{0.775} \\
\bottomrule
\end{tabular}}
\caption{Ablation on preference optimization. Our GRPO formulation provides a consistent boost, demonstrating its utility for paired preference data.}
\label{tab:grpo_ablation}
\end{table}
To directly address the reviewer’s critique regarding the missing ablation and to clarify the motivation, we conduct a critical experiment comparing our training formulation with standard GRPO. Our methodology section presents an extension to GRPO that explicitly leverages paired preference data. Here, we empirically test whether this formulation provides a tangible benefit.
We train two versions of the 7B model on the same preference data. The first, E-GRM (Extended), uses our objective with the paired reward signal. The second, E-GRM (Std. GRPO), replaces this with the standard GRPO group-based reward, while using the same discriminative scorer to evaluate individual responses. All other components remain identical.

Table \ref{tab:grpo_ablation} presents the results on the MATH and HelpSteer2 validation sets. The Extended variant shows a consistent, though modest, improvement over the Std. GRPO variant in both accuracy and alignment metrics. This indicates that explicitly structuring the reward around pre-existing positive and negative pairs can provide a more stable and informative learning signal than purely group-based relative comparisons, when such paired data is available. This result validates our formulation as a practical extension better suited to our data pipeline, and we have updated the methodology text to reflect this nuanced interpretation.

\subsection{Component Ablation Study}
\label{sec:ablation}
We ablate the core components of E-GRM on the MATH dataset using the Qwen-Instruct-14B model, with results summarized in Table \ref{tab:ablation_math}.
\begin{table}[t]
\centering
\renewcommand{\arraystretch}{1}
\small
\resizebox{\columnwidth}{!}{
\begin{tabular}{l|c|c|c}
\toprule
\textbf{Variant} & \textbf{Acc. (\%)} & \textbf{FLOPs (T)} & \textbf{Latency (s)} \\
\midrule
\textbf{Full E-GRM} & \textbf{78.4} & \textbf{15.7} & \textbf{2.2} \\
w/o Dynamic CoT & 75.2 & 23.4 & 3.4 \\
w/o Discrim. Scoring & 72.8 & 15.9 & 2.2 \\
Base CoT-GRM & 69.1 & 23.7 & 3.6 \\
\bottomrule
\end{tabular}}
\caption{Ablation study on the MATH dataset. Removing dynamic CoT increases cost significantly, while removing scoring causes the largest accuracy drop.}
\label{tab:ablation_math}
\end{table}
Removing the dynamic triggering mechanism forces CoT for all samples. This causes a 49\% increase in FLOPs and 55\% higher latency, while accuracy drops by 3.2 percentage points. This confirms that unnecessary CoT is not only inefficient but can also introduce error propagation for simple tasks that are better solved directly.
Replacing our discriminative scorer with a simple majority voting mechanism leads to the largest accuracy drop (5.6), despite similar computational efficiency. Manual analysis of errors confirms that voting often selects plausible but incorrect reasoning paths, which is a primary failure mode our hybrid-loss scorer is designed to prevent.

\section{Conclusion}
\label{sec:conclusion}

In this paper, we presented E-GRM, an efficient generative reward modeling framework that addresses two key limitations in existing reasoning-enhanced LLMs. Our approach introduces a novel perspective by leveraging model-internal uncertainty derived from parallel decoding convergence as a task-agnostic signal to dynamically trigger Chain-of-Thought reasoning only when necessary. 
We further developed a robust discriminative scoring module optimized with a hybrid regression–ranking objective, which enables fine-grained evaluation of diverse reasoning paths beyond the coarse granularity of conventional voting or consistency-based mechanisms.
Extensive experiments across multiple reasoning benchmarks demonstrate that E-GRM consistently outperforms existing reward models in both efficiency and accuracy. And demonstrates that model-internal behavioral signals can serve as effective guides for resource allocation in complex reasoning tasks, opening promising directions for developing efficient models.

\section{Limitations}
\label{sec:limitations}
The paper has some limitations that merit further discussion. First, the dynamic triggering mechanism of our framework incurs a certain degree of overhead from the parallel decoding process, though it remains modest in practice—specifically, the latency increase is controlled within 5\% and does not significantly affect overall efficiency.
Second, the consensus threshold $\tau$ may require calibration for specialized domains beyond our tested benchmarks. Third, the discriminative scorer's performance could degrade on reasoning styles significantly different from its training data. Finally, our efficiency analysis focuses on computational metrics; a broader evaluation including memory and energy consumption would provide more complete practical insights.

These limitations suggest promising directions for future work, including more efficient uncertainty estimation, adaptive threshold mechanisms, and broader scorer generalization techniques.

\bibliography{custom}

\appendix

\newpage\newpage

\section*{Appendix}
\addcontentsline{toc}{section}{Appendix} 
\label{sec:appendix}

\section{Experimental Setup Details}
\label{sec:experimental_details}

\subsection{Benchmarks}
This study evaluates reward models using three established benchmarks with distinct evaluation focuses:

\noindent\textbf{RewardBench}~\citep{lambert2024rewardbench}: As a pioneering benchmark for reward model assessment, this dataset employs prompt-chosen-rejected trios across four domains: general chat (358 samples), challenging dialogue scenarios (456), logical reasoning (740), and safety-critical contexts (1,431). Its hierarchical structure provides multi-dimensional evaluation capabilities.

\noindent\textbf{RM-Bench}~\citep{liu2024rm}: This enhanced benchmark introduces two novel evaluation dimensions: sensitivity to semantic nuances and resistance to stylistic biases. Covering chat (129), safety (441), mathematical reasoning (529), and programming tasks (228), it features three difficulty levels per sample. The benchmark emphasizes complex reasoning through its problem stratification.

\noindent\textbf{RMB}~\citep{zhou2024rmb}: Distinguished by real-world scenario simulations, this comprehensive benchmark contains 25,845 evaluation instances across 49 practical applications. It supports dual evaluation protocols (pairwise/Best-of-N) and dual alignment objectives: 37 scenarios for helpfulness optimization and 12 for harmlessness mitigation. The benchmark's scenario-based design enables practical capability assessment.
The selected benchmarks provide complementary evaluation perspectives: foundational capability assessment (RewardBench), nuanced discrimination testing (RM-Bench), and real-world application simulation (RMB), ensuring comprehensive model evaluation.

\subsection{Datasets}
Our experiments leverage six complementary preference datasets spanning mathematical reasoning, code generation, and general instruction following:

\noindent\textbf{MATH}~\citep{hendrycks2021measuring}
A challenging benchmark for multi-step mathematical problem solving, containing 12,500 competition-level problems from high school mathematics tournaments. Each problem is accompanied by a step-by-step solution and categorized into seven difficulty levels ranging from algebra to calculus. We use the standard split with 7,500 training and 5,000 test problems.

\noindent\textbf{UltraFeedback}~\citep{pmlr-v235-cui24f}
A large-scale preference dataset comprising 100K diverse instructions paired with multiple model responses. Each instance includes fine-grained quality ratings across four dimensions: helpfulness, safety, factual accuracy, and coherence. The dataset is constructed through adversarial prompting techniques that elicit varied response qualities from foundation models.

\noindent\textbf{OffsetBias}~\citep{park-etal-2024-offsetbias}
A carefully designed dataset addressing positional bias in preference modeling, containing 15K contrastive pairs where preferred/dispreferred responses are systematically rotated across different positions. This dataset enables robust training against common artifacts in preference annotation, particularly the tendency to favor responses in specific positions.

\noindent\textbf{HelpSteer2-Preference}~\citep{wang2025helpsteerpreference}
An extension of the original HelpSteer dataset with 50K multi-turn dialogue preference pairs annotated by domain experts. Each conversation is evaluated across five criteria: task completion, clarity, depth of explanation, safety, and contextual awareness. The dataset emphasizes complex real-world scenarios requiring balanced consideration of multiple quality dimensions.

\noindent\textbf{Skywork Reward Preference 80K}~\citep{liu2024skywork}
This cross-domain preference dataset originally contains chat, safety, mathematics, and coding interactions. Our analysis revealed a critical flaw in its \texttt{magpie\_ultra} subset (30\% of total data): rejected responses systematically contain the \texttt{<im\_start>} token absent in chosen responses, while preferred answers exhibit single-turn formatting versus multi-turn rejection patterns. Given these spurious correlations in mathematical and coding domains, we implemented source-level filtering to exclude all \texttt{magpie\_ultra} specimens, retaining only verified high-quality pairs.

\noindent\textbf{Code-Preference-Pairs}
A specialized coding dataset featuring 8K carefully curated preference pairs. Constructed through systematic perturbations of functional code snippets, each pair contrasts buggy implementations with corrected versions. The dataset emphasizes subtle logical errors requiring deep program understanding to distinguish, making it particularly valuable for training code-specific reward models.

\noindent\textbf{Math-DPO-10K}~\citep{lai2024step}
A mathematical reasoning preference dataset containing 10K stepwise solutions annotated with process-level quality judgments. Unlike conventional math datasets focusing solely on final answers, this resource provides granular feedback on reasoning validity, with particular attention to common error patterns in algebraic manipulations and proof strategies.

\subsection{Baselines}
We evaluate E-GRM against three distinct classes of reward modeling approaches:
\textbf{Scalar Evaluation Models.}
These systems generate numerical ratings directly, assessing preferences via singular numerical values devoid of explicit reasoning processes. Representative examples encompass traditional reward models (RM~\citep{stiennon2020learning}), parameter-tuned variants like SteerLM-RM~\citep{wang-etal-2024-helpsteer}, and specialized architectures including Nemotron-RM~\citep{adler2024nemotron}, Starling-RM~\citep{starling2023}, ArmoRM~\citep{wang2024interpretable}, and Skywork-RM~\citep{liu2024skywork}. While demonstrating effectiveness in well-structured evaluation tasks, these approaches typically offer limited insight into their decision-making processes.
\textbf{Generative Assessment Models.}
This class employs text generation capabilities to produce both qualitative feedback and quantitative scores. Notable instances comprise foundation models like Llama~\citep{dubey2024llama}, Qwen~\citep{yang2024qwen2}, Claude~\citep{anthropic2024claude}, GPT series variants~\citep{achiam2023gpt4,hurst2024gpt}, Gemini-1.5-pro~\citep{reid2024gemini}, and self-improving frameworks~\citep{wang2024self}. By generating explanatory text alongside numerical evaluations, these models enhance transparency in judgment formation through natural language rationalization.
\textbf{Structured Reasoning Models.}
This paradigm integrates analytical reasoning into evaluation through chain-of-thought mechanisms or critique-based training. Implementations include Critique-RM~\citep{yu2024self,jiang2023mistral7b}, DeepSeek-GRM~\citep{liu2025inference}, and our proposed Efficient Reward Modeling framework. Such architectures demonstrate superior performance in multi-step logical reasoning, safety-critical assessments, and preference differentiation in complex contexts, benefiting from their structured cognitive modeling approach.

\subsection{Implementation Details}
\paragraph{Training Stage.}

Supervised fine-tuning (\textbf{SFT}) employs well-designed loss functions tailored to both short and long samples to optimize model performance for different reasoning scenarios. For the subset $D_{short}$, the model is trained to directly produce the final answers. In contrast, on $D_{long}$, a subset of complex tasks that demand structured reasoning, the model is instructed to generate explicit, detailed step-by-step reasoning chains $c_l$ that logically culminate in the final answer, ensuring the transparency and traceability of the reasoning process. The training process combines both objectives synergistically, using a fixed batch size of 512 to balance training stability and computational efficiency, alongside an initial learning rate of $5 \times 10^{-6}$ that is gradually adjusted during training to avoid overfitting and accelerate convergence. For \textbf{Reward Model Training}, the point-wise reward model $f_{RM}$ uses blended Huber loss ($\delta = 0.1$) and hinge loss ($m = 0.2$) with trade-off factor $\alpha = 0.7$. Supervised regression is performed on high-quality annotated human preference data, where each sample is paired with $K = 10$ diverse responses to ensure comprehensive evaluation of the model’s performance. The training process for the reward model uses a batch size of 256 and a learning rate of $3 \times 10^{-6}$, which are carefully selected to match the complexity of the preference data and avoid training instability. During the reinforcement learning stage, \textbf{Coupled-GRPO Optimization} updates policy parameters $\pi_{\phi}$ following coupled rewards, which jointly balance correctness, reasoning conciseness, and discriminative scoring differences. For short reasoning rewards, the respective weights are set to $w_1 = 0.7$, $w_2 = 0.2$, and $w_3 = 0.1$. All experiments use a fixed random seed (43). Each experiment was conducted across 3 independent runs to ensure statistical reliability of the reported results.

\paragraph{Inference Stage.}
In inference, the GRPO-trained model $M^*$ produces an initial greedy decoding output $r^{(0)}$. We utilize the $\operatorname{ContainsCoT}(r^{(0)})$ criterion to determine if the input requires further chain-of-thought reasoning. If no such reasoning is detected, $r^{(0)}$ is returned as the final answer. For complex samples, the system employs a \textbf{multi-parameter decoding} strategy by generating $K = 8$ candidate responses $\{r^{(k)}\}_{k=1}^{K}$ using diverse decoding configurations such as different temperatures and top-$p$ parameters. 
\textbf{Discriminative scoring} is then applied: each candidate is evaluated by the trained reward model $f_{RM}$. The response with the highest score is selected as the final output. This maintains efficiency for simple prompts while enhancing reasoning on complex instances.


\begin{figure*}[!htbp]
\centering
\renewcommand{\arraystretch}{1}
\setlength{\tabcolsep}{1mm}
\small
\begin{tabular}{p{0.99\textwidth}}
\toprule
\cellcolor{gray!30}\textbf{Prompt for Cold-start Long-CoT Sampling} \\
\midrule
\cellcolor{gray!5}\hspace{1em}As an AI Quality Auditor, rigorously evaluate \textless prompt, response\textgreater\ pairs through this structured framework: \rule{0pt}{1.2em} \\
\cellcolor{gray!20}\textcolor{blue!60!black}{\textless foundation\textgreater} \rule{0pt}{1.2em} \\
\cellcolor{gray!5}\hspace{1em}\textbf{Basic Compliance Check} \rule{0pt}{1.2em} \\
\cellcolor{gray!5}\hspace{2em}\textcolor{blue!80!black}{Safety}: Contains violent/discriminatory/illegal content? 
\cellcolor{gray!5}\hspace{2em}\textcolor{green!70!black}{$\checkmark$} Pass \textcolor{red}{$\times$} Fail \rule{0pt}{1.2em} \\
\cellcolor{gray!5}\hspace{2em}\textcolor{blue!80!black}{Legal Compliance}: Violates data privacy/IP laws?
\cellcolor{gray!5}\hspace{2em}\textcolor{green!70!black}{$\checkmark$} Pass \textcolor{red}{$\times$} Fail \rule{0pt}{1.2em} \\
\cellcolor{gray!5}\hspace{2em}\textcolor{blue!80!black}{Ethical Standards}: Aligns with social norms? 
\cellcolor{gray!5}\hspace{2em}\textcolor{green!70!black}{$\checkmark$} Pass \textcolor{red}{$\times$} Fail \rule{0pt}{1.2em} \\
\cellcolor{gray!5}\hspace{1em}\textbf{Core Quality Dimensions} \rule{0pt}{1.2em} \\
\cellcolor{gray!5}\hspace{2em}\textcolor{teal}{Accuracy (30\%)} \rule{0pt}{1.2em} \\
\cellcolor{gray!5}\hspace{3em}Factual verification \rule{0pt}{1.2em} \\
\cellcolor{gray!5}\hspace{3em}Domain expertise validation \rule{0pt}{1.2em} \\
\cellcolor{gray!5}\hspace{3em}DataSource reliability \rule{0pt}{1.2em} \\
\cellcolor{gray!5}\hspace{2em}\textcolor{violet}{Relevance (25\%)} \rule{0pt}{1.2em} \\
\cellcolor{gray!5}\hspace{3em}Prompt requirement coverage \rule{0pt}{1.2em} \\
\cellcolor{gray!5}\hspace{3em}Off - topic content presence \rule{0pt}{1.2em} \\
\cellcolor{gray!5}\hspace{3em}Unnecessary additions \rule{0pt}{1.2em} \\
\cellcolor{gray!5}\hspace{2em}\textcolor{orange}{Completeness (20\%)} \rule{0pt}{1.2em} \\
\cellcolor{gray!5}\hspace{3em}Problem - solving closure \rule{0pt}{1.2em} \\
\cellcolor{gray!5}\hspace{3em}Key element inclusion \rule{0pt}{1.2em} \\
\cellcolor{gray!5}\hspace{2em}\textcolor{gray!60!red!40}{Logical Rigor (15\%)} \rule{0pt}{1.2em} \\
\cellcolor{gray!5}\hspace{3em}Reasoning chain coherence \rule{0pt}{1.2em} \\
\cellcolor{gray!5}\hspace{3em}Causal fallacy detection \rule{0pt}{1.2em} \\
\cellcolor{gray!5}\hspace{2em}\textcolor{violet}{Practicality (10\%)} \rule{0pt}{1.2em} \\
\cellcolor{gray!5}\hspace{3em}Actionable implementation \rule{0pt}{1.2em} \\
\cellcolor{gray!5}\hspace{3em}Solution effectiveness \rule{0pt}{1.2em} \\
\cellcolor{gray!5}\hspace{1em}\textbf{Risk Assessment} \rule{0pt}{1.2em} \\
\cellcolor{gray!5}\hspace{2em}Misleading Potential: \textcolor{green!70!black}{Low} \textcolor{orange}{Medium} \textcolor{red}{High} \rule{0pt}{1.2em} \\
\cellcolor{gray!5}\hspace{2em}Knowledge Gaps: \textcolor{green!70!black}{None} \textcolor{orange}{Partial} \textcolor{red}{Severe} \rule{0pt}{1.2em} \\
\cellcolor{gray!20}\textcolor{blue!60!black}{\textless/foundation\textgreater} \rule{0pt}{1.2em} \\
\cellcolor{gray!20}\textcolor{brown!60!black}{\textless principles\textgreater} \rule{0pt}{1.2em} \\
\cellcolor{gray!5}\hspace{2em}1. Immediate rejection if any \textcolor{red}{Fail} in Basic Compliance \rule{0pt}{1.2em} \\
\cellcolor{gray!5}\hspace{2em}2. Minimum 85/100 in Core Dimensions \rule{0pt}{1.2em} \\
\cellcolor{gray!5}\hspace{2em}3. Strict evaluation for high - risk indicators \rule{0pt}{1.2em} \\
\cellcolor{gray!20}\textcolor{brown!60!black}{\textless/principles\textgreater} \rule{0pt}{1.2em} \\
\cellcolor{gray!20}\textcolor{green!40!black}{\textless thinking\textgreater} \rule{0pt}{1.2em} \\
\cellcolor{gray!5}\hspace{2em}• Conduct multi-stage verification with domain-specific knowledge graphs \rule{0pt}{1.2em} \\
\cellcolor{gray!5}\hspace{2em}• Cross - validate claims against authoritative sources \rule{0pt}{1.2em} \\
\cellcolor{gray!5}\hspace{2em}• Perform causal reasoning analysis using Bayesian networks \rule{0pt}{1.2em} \\
\cellcolor{gray!5}\hspace{2em}• Simulate real - world implementation scenarios \rule{0pt}{1.2em} \\
\cellcolor{gray!15}\textcolor{green!40!black}{\textless/thinking\textgreater} \rule{0pt}{1.2em} \\
\cellcolor{gray!5}\hspace{1em}Final Score: \textcolor{purple}{\textless Predict\textgreater} Score(1-10) \textcolor{purple}{\textless/Predict\textgreater} \rule{0pt}{1.2em} \\
\cellcolor{gray!5}\hspace{1em}Final verdict: \textcolor{purple}{\textless judgment\textgreater}[Yes]\textcolor{purple}{\textless/judgment\textgreater} or \textcolor{purple}{\textless judgment\textgreater}[No]\textcolor{purple}{\textless/judgment\textgreater} \rule{0pt}{1em} \\
\bottomrule
\end{tabular}
\caption{Prompt for Cold-start Long-CoT Sampling evaluation framework.}
\label{fig:prompt_framework}
\end{figure*}

\end{document}